\documentclass[10pt,twocolumn,letterpaper]{article}

\usepackage{cvpr}              

\usepackage{graphicx}
\usepackage{amsmath}
\usepackage{amssymb}

\usepackage{booktabs}
\usepackage{array}
\usepackage{multirow}
\usepackage[pagebackref,breaklinks,colorlinks]{hyperref}

\usepackage[capitalize]{cleveref}
\crefname{section}{Sec.}{Secs.}
\Crefname{section}{Section}{Sections}
\Crefname{table}{Table}{Tables}
\crefname{table}{Tab.}{Tabs.}
\usepackage{color}

\def\cvprPaperID{4650} 
\def\confName{CVPR}
\def\confYear{2023}

\begin{document}

\title{Boosting Weakly-Supervised Temporal Action Localization with Text Information}

\author{Guozhang Li$^1$,De Cheng$^1$,Xinpeng Ding$^2$, Nannan Wang$^1$\thanks{Corresponding author} , Xiaoyu Wang$^3$, Xinbo Gao$^{1,4}$\\
{$^1$}Xidian University,{$^2$}The Hong
Kong University of Science and Technology, \\{$^3$}The Chinese University of Hong Kong (Shenzhen)
{$^4$}Chongqing University of Posts and Telecommunications\\
{\tt\small liguozhang@stu.xidian.edu.cn, dcheng@xidian.edu.cn, xdingaf@connect.ust.hk}\\
{\tt\small nnwang@xidian.edu.cn,fanghuaxue@gmail.com, gaoxb@cqupt.edu.cn}
}
\maketitle


\maketitle


\begin{abstract}
Due to the lack of temporal annotation, current Weakly-supervised Temporal Action Localization (WTAL) methods are generally stuck into over-complete or incomplete localization.
%
In this paper, we aim to leverage the text information to boost WTAL from two aspects, ~\ie, \textbf{(a)} the discriminative objective to enlarge the inter-class difference, thus reducing the over-complete;
\textbf{(b)} the generative objective to enhance the intra-class integrity, thus finding more complete temporal boundaries.
%
For the discriminative objective, we propose a Text-Segment Mining (TSM) mechanism, which constructs a text description based on the action class label, and regards the text as the query to mine all class-related segments.
%
%
%
Without the temporal annotation of actions, TSM compares the text query with the entire videos across the dataset to mine the best matching segments while ignoring irrelevant ones.
%
%
Due to the shared sub-actions in different categories of videos, merely applying TSM is too strict to neglect the semantic-related segments, which results in incomplete localization.
We further introduce a generative objective named Video-text Language Completion (VLC), which focuses on all semantic-related segments from videos to complete the text sentence.
%
%
We achieve the state-of-the-art performance on THUMOS14 and ActivityNet1.3.
%
Surprisingly, we also find our proposed method can be seamlessly applied to existing methods, and improve their performances with a clear margin. The code is available at \href{https://github.com/lgzlIlIlI/Boosting-WTAL}{https://github.com/lgzlIlIlI/Boosting-WTAL}.
%
%
%

%
%
%
%

\if
\fi
\end{abstract}
\begin{figure}[!tbp]
\begin{center}
    \includegraphics[width=1.0\linewidth]{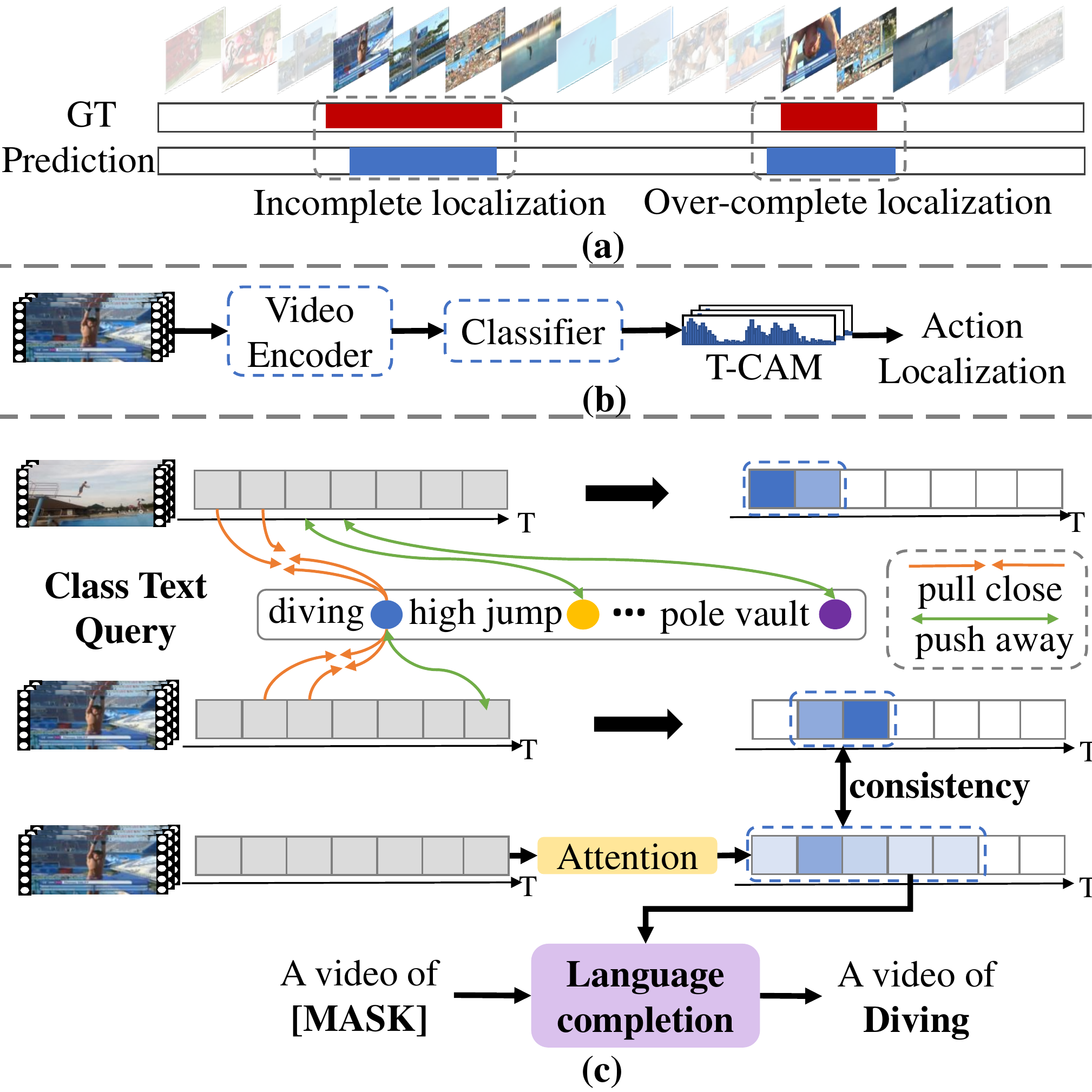}
    \vspace{-0.8cm}
\end{center}
   \caption{{Comparison of our proposed framework with current WTAL methods. (a) Common failures in existing WTAL methods. (b) Existing WTAL model's pipeline. (c) The proposed framework with text-segment mining and video-text language completion, where the depth of color represents the degree of correlation between segments and texts. }}
\label{fig:mot}
\vspace{-0.5cm}
\end{figure}
\section{Introduction}
\label{sec:intro}

Temporal action localization attempts to temporally localize the action instances of interest in untrimmed videos.
Although current fully-supervised temporal action localization methods \cite{Lin_2019_ICCV,tan2021relaxed,ding2021crnet,zhang2022actionformer} have achieved remarkable progress, time-consuming and labor-intensive frame-level annotations are required.
To alleviate the annotation cost, weakly-supervised temporal action localization (WTAL) methods \cite{paul2018w,lee2019background,narayan2021d2,huang2022weakly}
have gained more attention recently, which only requires efficient video-level annotations.

With only video-level supervision, existing WTAL methods~\cite{wang2017untrimmednets,paul2018w,lee2019background,huang2022weakly} generally utilize video information to train a classifier, which is used to generate a sequence of class logits or predictions named temporal class activation map (T-CAM).
%
%
%
%
%
%
%
While significant improvement has been achieved, current methods still suffer from two problems,~\emph{i.e.},~incomplete and over-complete localization.
As shown in Fig.~\ref{fig:mot} (a), some sub-action with low discriminability may be ignored, while some background segments that contribute to classification can be misclassified as action, causing incomplete and over-complete localization.
Differently from current methods that only utilize the video information, in this paper, we aim to explore the text information to improve WTAL from two aspects: \textbf{(a)} the discriminative objective to enlarge the inter-class difference, thus reducing the over-complete;
\textbf{(b)} the generative objective to enhance the intra-class integrity, thus finding more complete temporal boundaries.
%
%
%
For the discriminative objective, we propose a Text-Segment Mining (TSM) mechanism, where the action label texts can be used as queries to mine all related segments in videos. 
Specifically, we first use the prompt templates to incorporate the class label information into the text query.
%
Without temporal annotations, TSM requires to compare the text query with all segments of the different videos across the dataset, as shown in Fig.~\ref{fig:mot} (c).
During the comparison, the segments that is best matching to the text query would be mined, while other irrelevant segments would be ignored, which is similar to `matched filter'~\cite{turin1960introduction,Zhang_2021_CVPR}.
In this way, the segments and text queries with the same class from all videos would be pulled close while pushing away others, hence enhancing the inter-class difference.
%
%
%

\begin{figure*}[!t]
\setlength{\abovecaptionskip}{0.cm}
\setlength{\belowcaptionskip}{-0.cm}
\begin{center}
    \includegraphics[width=1.0\linewidth]{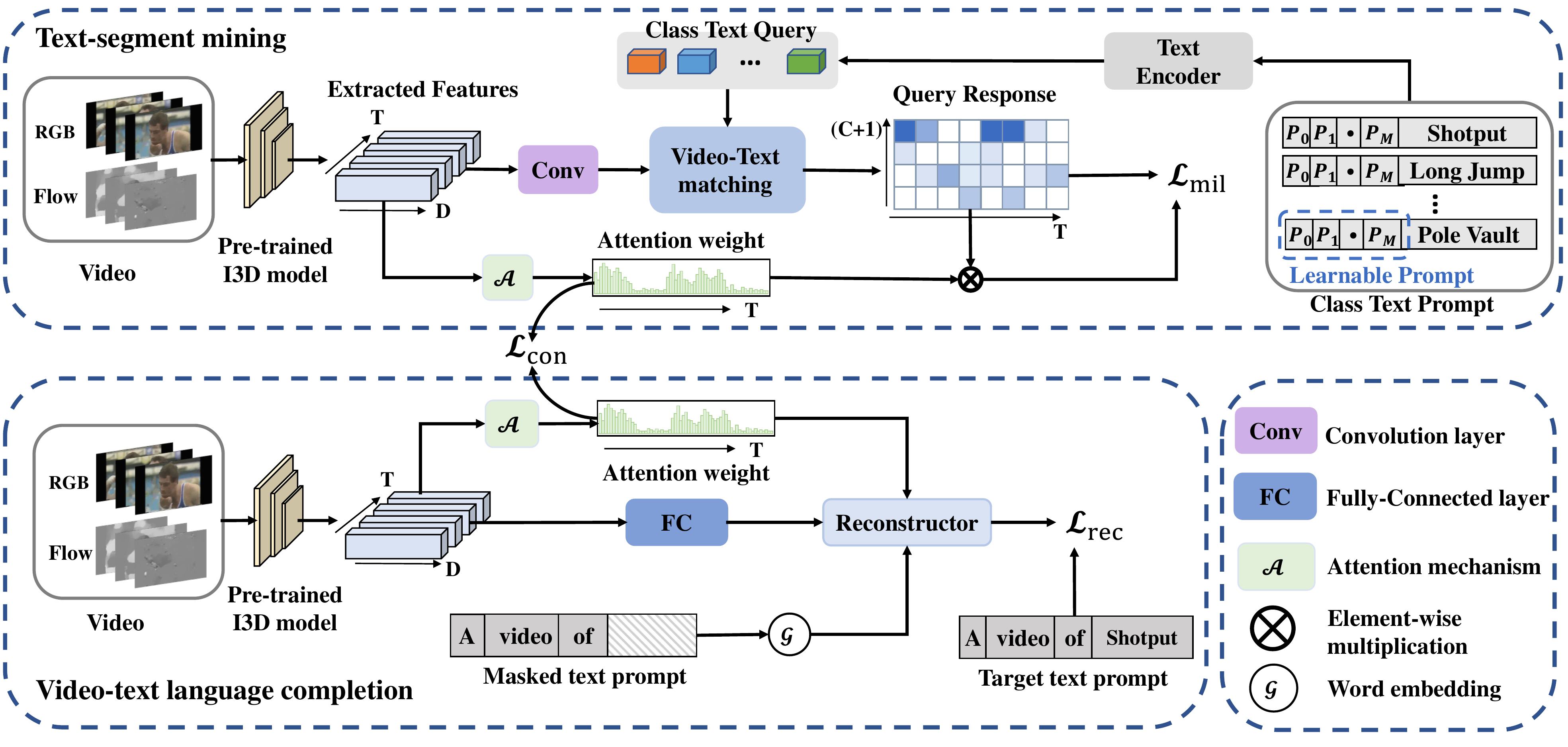}
    \vspace{-0.5cm}
\end{center}
   \caption{{Illustration of the proposed framework. In this work, the text-segment mining objective uses the action label texts as a query to mine semantically related segments in the video to achieve action localization. In addition, the language completion objective aims to focus on the areas related to the action label texts in the video as comprehensively as possible to complete the masked keywords, and alleviate the localization errors caused by the excessive attention of the matching strategy to the most relevant segments in a self-supervised manner. }}
\label{fig:fra}
\vspace{-0.5cm}
\end{figure*}

For different categories of videos, there are some shared sub-actions,~\emph{e.g.}, sub-action ``Approach'' exists in both ``High Jump'' and ``Long Jump'' videos.
Merely using TSM is too strict to neglect the semantic-related segments, which results in incomplete localization,~\emph{e.g.}, neglecting ``Approach'' segments.
%
%
To overcome this problem, we further introduce a generative objective named Video-text Language Completion (VLC) which focuses on all semantic-related segments to complete the text sentence.
First, we construct a description sentence for the action label of the video and mask the key action words in the sentence, as shown in Fig.~\ref{fig:fra}.
Then an attention mechanism is design to collect semantic related segments as completely as possible to predict masked action text via the language reconstructor, which enhances the intra-class integrity.
%
%
%
%
%
Combining TSM and VLC by a self-supervised constraint, our method achieves the new state-of-the-art on two popular benchmarks,~\emph{i.e.}, THUMOS14~\cite{THUMOS14} and ActivityNet1.3~\cite{caba2015activitynet}.
%
%
Furthermore, we also find our proposed method can be applied into existing methods, and improve their performances with a clear margin.

Our contributions are summarized as three-folds:
%
%
\textbf{(a)} To best of our knowledge, we are the first to leverage text information to boost WTAL. We also prove that our method can be easy to extend to existing state-of-the-art approaches and improve their performance. 
%
\textbf{(b)} To leverage the text information, we devise two objective: the discriminative objective to enlarge the inter-class difference, thus reducing the over-complete; and the generative objective to enhance the intra-class integrity, thus finding more complete temporal boundaries.
%
\textbf{(c)} Extensive experiments illustrate our method outperforms current methods on two public datasets, and comprehensive ablation studies reveal the effectiveness of the proposed objectives. 

\section{Related Work}
\label{sec:related}

\noindent \textbf{Weakly Supervised Temporal Action Localization.}
Weakly-supervised temporal action localization requires video-level labels only. 
Due to the lack of precise boundary labels, most advanced WTAL methods~\cite{paul2018w,min2020adversarial,islam2021hybrid,huang2022weakly} fall into a localization-by-classification pipeline to tackle the WTAL task. 
Erasing-based methods~\cite{singh2017hide,zeng2019breaking,zhong2018step,min2020adversarial} carefully design adversarial erase strategies, which find many less discriminative regions by erasing the most discriminant regions. 
Metric learning-based methods~\cite{paul2018w,huang2020relational,narayan20193c,2020Adversarial} employ center loss or triple loss to decrease intra-class variations while increasing inter-class difference.
In addition, background segments suppression-based methods~\cite{lee2019background,lee2021weakly,islam2021hybrid} aim to separate action segments from background segments by setting additional background class to learn background suppression weights.
Some pseudo-label-based methods \cite{Zhai2020TwoStreamConsensus,2020WeaklyEM,huang2022weakly} utilize video information to generate pseudo-labels to improve the quality of T-CAMs. Besides, lee ~\emph{et al}. \cite{lee2021cross} used audio within the video as an auxiliary information. 
Existing methods can employ one or more of the above strategies to improve T-CAM quality and improve localization performance. 
Despite the success of these methods, however, the above strategies only make use of video information, and the semantic information encapsulated in category labels of the text form is not fully explored. 
In this paper, we design a novel framework consisting of two objectives, \emph{i.e.,} text-segment mining and video-text language completion, to leverage action label text information to boost WTAL.

\noindent \textbf{Self-Supervised Learning.}
Self-supervised learning leverages unlabeled data to make the model learn intrinsic information from data.
Currently, several methods have been proposed to exploit the self-supervised learning strategy to learn better representation when lacking full annotation data.
For example, Gong \emph{et.al} \cite{gongself} proposed self-supervised equivariant transform consistency constraint to realize self-supervised action localization.
TSCN \cite{Zhai2020TwoStreamConsensus} and UGCT \cite{yang2021uncertainty} utilize RGB and optical flow video features for cross-supervision to improve the performance of WTAL.
Su \emph{et.al} \cite{su2021improving} utilizes temporal multi-resolution information to generate pseudo labels for better representation learning.
VLC model tends to focus on all video segments related to action text to achieve text integrity, which can be used to alleviate excessive attention to important segments in the TSM model.
In this paper, we utilize the label text information to construct a VLC model and design a self-supervised constraint between the TSM and the VLC model to achieve more complete localization results.

\noindent \textbf{Vision-Language Models.}
Recently, a series of works on the interaction of vision and language has attracted increasing attention in the past few years, such as vision language pre-training \cite{radford2021learning,ju2021prompting}, video caption \cite{wang2018reconstruction}, video grounding \cite{Ding_2021_ICCV,zhang2020learning,mun2020local}, video question answering \cite{lei2019tvqa+} and so on.
However, how to make full use of the information encapsulated in action label texts in the WTAL task has not yet been explored. 
In this paper, we design a novel framework to explore leverage text information of  action label to boost WTAL task.
By combining discriminative objective TSM and the generative objective VLC, the proposed framework realizes the indirect use of text information to boost WTAL.


\section{The Proposed Method}
\label{sec:related}

\subsection{Overall Architecture}

\noindent \textbf{Problem formulation.} In WTAL, we are provided with a set of $N$ untrimmed videos defined as $\{{V_j}\}_{j=1}^N$, and all of them are annotated with their corresponding video-level action category labels $\{\mathbf{y}_j\}_{j=1}^N$. Generally, the label $\mathbf{y}_j$ is discretized into a binary vector indicating the presence/absence of each category of action in the video $v_j$. Each video $V$ contains a set of segments: ${V} = \{ {v_t}\}_{t=1}^{\mathcal{T}}$, where $\mathcal{T}$ is the number of segments in the video. Generally, $\mathcal{T}$ segments are fed into a pre-trained 3D CNN model \cite{DBLP:conf/cvpr/CarreiraZ17} to extract both RGB features $\mathbf{X}_r \in \mathbb{R}^{\mathcal{T}\times 1024}$ and FLow video features $\mathbf{X}_f \in \mathbb{R}^{\mathcal{T}\times 1024}$. {During inference, we predict a sequence of actions $\{c_i, s_i, e_i, {conf_i}\}$ for an input video, where $c_i$ is the action category, $s_i$ and $e_i$ represent the start and end time, and ${conf_i}$ is confidence score.}

\noindent \textbf{Overview.}
The proposed overall framework is shown in Figure \ref{fig:fra}, which leverages text information of action labels to boost WTAL from two aspects, \ie, Text-Segment Mining (TSM) and Video-text Language Completion (VLC).
For the TSM in Section \ref{TSM}, the RGB and Flow video features $\mathbf{X}_r$ and $\mathbf{X}_f$ is fed into a video embedding module consisting of convolution layers to generate video features embedding at first.
Second, we construct text descriptions for action labels via prompt template and generate text queries according to the description by the text encoder.
Then in the video-text matching module, TSM compares the text queries with all segments of videos to generate query responses to mine semantically related video segments. 
In addition, we generate attention weights for each video segment to further suppress the response of the background segments to text queries.
For the VLC in Section \ref{SC}, the extracted video features $\mathbf{X}_r$ and $\mathbf{X}_f$ are fed into a fully connected layer to get video feature embedding at first.
Later, we construct a description sentence for the action label of the video and mask the key action words of the sentence.
Then, an attention mechanism is designed to collect semantically related segments to predict masked action words via the language reconstructor.
%
%
%
%
%
%
%
%
%
%
Finally, in Section \ref{SCC}, we combine the TSM and VLC via imposing self-supervised constraints between attentions of them to obtain more accurate and complete localization results.
\subsection{Text-Segment Mining}
\label{TSM}
In this section, we introduce the text-segment mining objective (TSM) to make full use of information encapsulated in action label texts. 
Specifically, the TSM consists of a video embedding module, a text embedding module and a video-text feature matching module. 

\noindent \textbf{Video embedding module.} Similar to other WTAL models, the video embedding module is composed of two 1D convolutions followed by RelU and Dropout layers.
We use a strategy similar to \cite{hong2021cross} to fuse RGB and Flow features to obtain video features $\mathbf{X} \in \mathbb{R}^{\mathcal{T}\times 2048}$ as the input of the video embedding module.
Then, the corresponding video feature embedding $\mathbf{X}_e \in \mathbb{R}^{\mathcal{T}\times 2048}$ can be obtained by $\mathbf{X}_e = emb(\mathbf{X})$, where $emb(\cdot)$ represents the video embedding module.
Besides, following previous works \cite{islam2021hybrid,hong2021cross}, an attention mechanism is utilized to generate attention weight $\mathbf{{att}}_{m} \in \mathbb{R}^{\mathcal{T}\times 1}$ for each video segment $V_j$,
\begin{equation} 
\label{E4} 
{
\mathbf{{att}}_{m} = \sigma(\mathcal{A}(\mathbf{{X}})),
}
\end{equation}
where $\mathcal{A}(\cdot)$ is the attention mechanism consisting of several convolution layers, and $\sigma(\cdot)$ means the sigmoid function.

\noindent \textbf{Text embedding module.} The text embedding module aims to use action label text to generate a series of queries for mining segments related to category text in the videos. 
We adopt category-specific learnable prompts for $C$ category of action label texts, to form the input of the text embedding module $L_q$:
\begin{equation} 
\begin{aligned}
\label{E6} 
{
\mathbf{L}_{q} = [\mathbf{L}_{s};\mathbf{L}_{p};\mathbf{L}_{e}],
}
\end{aligned}
\end{equation}
where $\mathbf{L}_{s}$ denotes the [START] token initialized randomly, $\mathbf{L}_{p}$ denotes the learnable textual contexts with the length $N_p$, and $\mathbf{L}_{e}$ denotes action label text features embed by GloVe \cite{pennington2014glove}.
Besides, the $C+1$-th additional background class embedding is initialized by zero.

Then a Transformer encoder $trans(\cdot)$ is used as the text embedding module to generate text queries.
Specifically, the class text queries $\mathbf{X}_{q}$ can be obtained by $\mathbf{X}_{q} = trans(\mathbf{L}_{q})$, where $\mathbf{X}_q \in \mathbb{R}^{(C+1) \times 2048}$.

\noindent \textbf{Video-text feature matching.} The video-text feature matching module is used to match semantic-related text query and video segment features. 

To be specific, we conduct the inner product operation between the video embedding feature $\mathbf{X}_e$ and the text queries $\mathbf{X}_{q}$ to generate the segment-level video-text similarity matrix $\mathbf{S} \in \mathbb{R}^{\mathcal{T}\times (C+1)}$.

Besides, following the background suppression-based methods \cite{lee2019background,islam2021hybrid,hong2021cross}, we also apply attention weight $\mathbf{att}_{m}$ to suppress the response of the background segment to the action text. 
The background suppressed segment-level matching result $\mathbf{\bar{S}} \in \mathbb{R}^{\mathcal{T}\times C+1}$ can be obtained by $\mathbf{\bar{S}} = \mathbf{att}_{m} * \mathbf{S}$, where `*' means element-wise multiplication in this paper.


Finally, similar to current approaches \cite{paul2018w,2020Adversarial}, We also use top-k multi-instance learning to calculate matching loss.
Specifically, we calculate the average value of top-$k$ similarity in the temporal dimension corresponding to a specific category of text query as the video-level video-text similarity.

For the $j$-th action category, video-level similarity $\mathbf{v}_{j}$ and $\bar{\mathbf{v}}_{j}$ are generated from $\mathbf{S}$ and $\mathbf{\bar{S}}$, respectively:
\begin{equation}
\label{E10} 
\mathbf{v}_{j}=\max _{\substack{l \subset\{1, \ldots, \mathcal{T}\} \\|l|=k}} \frac{1}{k} \sum_{i \in l} \mathbf{S}_{i}(j), 
\quad
\bar{\mathbf{v}}_{j}=\max _{\substack{l \subset\{1, \ldots, \mathcal{T}\} \\|l|=k}} \frac{1}{k} \sum_{i \in l} \mathbf{\bar{S}}_{i}(j),
\end{equation}
where $l$ is a set containing the index of the top-$k$ segments with the highest similarity to the j-$th$ text query, and $k$ is the number of selected segments.
Then, We apply softmax to  $\mathbf{v}_{j}$ and $\bar{\mathbf{v}}_{j}$ to generate video-level similarity score $\mathbf{p}_{j}$ and $\bar{\mathbf{p}}_{j}$.

We encourage the positive score of video-text category matching to approach 1, while the negative score to approach zero to train the TSM objective,

\begin{equation}
\label{E12}
\mathcal{L}_{mil}=-(\sum_{j=1}^{C+1} \mathbf{y}_{j} \log \left(\mathbf{p}_{j}\right)+\sum_{j=1}^{C+1} \hat{\mathbf{y}}_{j} \log \left(\hat{\mathbf{p}}_{j}\right)),
\end{equation}
where $\mathbf{y}_{j}$ and $\hat{\mathbf{y}}_{j}$ are labels for video-text matching. In addition, the additional $C+1$-th background class is 0 in $\hat{\mathbf{y}}_{j}$ and 1 in $\mathbf{y}_{j}$. 

Besides, in this work, follow \cite{islam2021hybrid,hong2021cross}, we also adopt co-activity loss \cite{paul2018w,2020Adversarial}, normalization loss \cite{lee2019background,lee2021weakly} and guide loss \cite{islam2021hybrid,hong2021cross} to train the TSM model.
Since they are not the main contributions of this work, we do not elaborate on them in this paper. 

\subsection{Video-Text Language Completion}
\label{SC}
The Video-text Language Completion (VLC) objective aims to complete the masked keywords in the video description, by focusing on the text-related video segments related as comprehensively as possible.
The proposed VLC also contains a video embedding module and a text embedding module. 
Besides, a transformer reconstructor is used for multi-modal interaction and completion of the original text description.    

\noindent \textbf{Video embedding module.} 
%
Given the original video feature $\mathbf{X} \in \mathbb{R}^{\mathcal{T}\times 2048}$ as described in sec.\ref{TSM}, we can obtain the corresponding video feature embedding  $\mathbf{X}_v \in \mathbb{R}^{\mathcal{T}\times 512}$ through for a full connection layer the VLC module.

To mine positive areas of text-semantic-related video,
the proposed completion model specially designs an attention mechanism with the same structure as Sec. \ref{TSM}. 
The attention weight for VLC $\mathbf{{att}}_{r} \in \mathbb{R}^{\mathcal{T}\times 1}$ can be obtained by:
\begin{equation} 
\label{E44} 
{
\mathbf{{att}}_{r} = \sigma(\mathcal{A}(\mathbf{{X}})),
}
\end{equation}
where $\mathcal{A}(\cdot)$ is the attention mechanism composed of several convolution layers, and $\sigma(\cdot)$ represents the sigmoid function.

\noindent \textbf{Text embedding module.} The datasets of the WTAL task only provide action videos and their action labels but does not contain any sentences describing the corresponding videos. 
Hence, we first use the prompt template ``a video of [CLS]'' and the action label texts to construct a description sentence for the video.
Then, we mask the key action words of the description sentence, and embed the masked sentence with GloVe \cite{pennington2014glove} and a fully connected layer to get sentence feature embedding $\hat{\mathbf{X}}_s \in \mathbb{R}^{M \times 512}$, where $M$ is the length of the sentence.

\noindent \textbf{Transformer reconstructor.} In the video-text language completion model, a transformer reconstructor is used to complete the masked description sentence.
Firstly, following \cite{lin2020weakly}, we randomly mask 1/3 of the words in the sentence as the alternative description sentence, which could result in a high probability to mask the action label texts. 
%
Then, the encoder of the transformer is used to get the foreground video feature $\mathbf{F} \in \mathbb{R} ^ {\mathcal{T}\times 512}$ by:
\begin{equation}
\label{E11}
\mathbf{F} = E(\mathbf{X}_v,\mathbf{att}_{r}) = \delta(\frac{\mathbf{X}_v\mathbf{W}_q(\mathbf{X}_v\mathbf{W}_k)^T}{\sqrt{D_h}}*\mathbf{att}_{r})\mathbf{X}_v\mathbf{W}_v,
\end{equation}
where $E(\cdot,\cdot)$ denotes the transformer encoder, $\delta(\cdot)$ denotes the softmax function, $\mathbf{W}_q, \mathbf{W}_k,\mathbf{W}_v \in \mathbb{R}^{512 \times 512}$ are learnable parameters, and $D_h = 512$ is the feature dimension of $\mathbf{X}_v$.
The decoder of the transformer is used to obtain multi-modal representation $\mathbf{H} \in \mathbb{R}^{M \times 512}$ to reconstruct the masked sentence:
\begin{equation}
\label{E12}
\begin{split}   \mathbf{H}&=D(\hat{\mathbf{X}}_s,\mathbf{F},\mathbf{att}_{r}) \\ &=\delta(\frac{\mathbf{\hat{X}}_s\mathbf{W}_{qd}(\mathbf{F}\mathbf{W}_{kd})^T}{\sqrt{D_h}}*\mathbf{att}_{r})\mathbf{F}\mathbf{W}_{vd},\end{split}
\end{equation}
where $D(\cdot,\cdot,\cdot)$ denotes the transformer decoder, and $W_{qd},W_{kd},W_{vd} \in \mathbb{R}^{512 \times 512}$ are learnable parameters.

Finally, the probability distribution $\mathbf{P} \in \mathbb{R}^{M\times N_v}$ of the $i$-th word $w_i$ on the vocabulary can be obtained by:
\begin{equation}
\label{E13}
\mathbf{P}(w_i|\mathbf{X}_{v},\mathbf{\hat{X}}_{s[0:i-1]}) = \delta(FC(\mathbf{H})),
\end{equation}
where FC($\cdot$) denotes the fully connected layer, $\delta(\cdot)$ denotes the softmax function, and $N_v$ is the vocabulary size.

The final VLC loss function can be formulated as:
\begin{equation}
\label{E14}
\mathcal{L}_{rec} = -\sum^{M}_{i=1} \log\mathbf{P}(w_i|\mathbf{X}_{v},\mathbf{\hat{X}}_{txt[0:i-1]})).
\end{equation}

To further improve the mined positive areas of text-semantic-related video, we also adopt a contrastive loss \cite{zheng2022weakly} in the completion model.
Specifically, positive areas mined by attention weight $\mathbf{att}_{r}$ should be more compatible with the sentence than the entire video, and those negative areas mined by $1-\mathbf{att}_{r}$.
Therefore, following Eq. \ref{E11}-\ref{E14}, we can obtain the completion loss $\mathcal{L}_{rec}^{e}$ and $\mathcal{L}_{rec}^{n}$, where the attention weight $\mathbf{att}_{r}$ used in the transformer is replaced with $1$ and $1-\mathbf{att}_{r}$, respectively.

Finally, the contrastive loss $\mathcal{L}_{c}$ can be formulated as:
\begin{equation}
\label{E15}
\mathcal{L}_{c} = \max(\mathcal{L}_{rec}-\mathcal{L}_{rec}^{e} + \gamma_1,0) + \max(\mathcal{L}_{rec}-\mathcal{L}_{rec}^{n} + \gamma_2,0),
\end{equation}
where $\gamma_1$ and $\gamma_2$ are hyper-parameters.

\subsection{Self-Supervised Consistency Constraint}
\label{SCC}
The matching strategy used in TSM tends to focus on the video segments that better match the text, while excluding other text-unrelated segments as they could lead to localization error. 
%
On the other hand, the VLC tends to focus on all video clips that are related to action text to achieve description completion.
Hence, we impose self-supervised constraints between attentions of these two objectives, \emph{i.e.,} the discriminative objective TSM and the generative objective VLC, to alleviate the excessive attention paid to the most semantic-related segments by TSM.
The consistency constraint loss $L_{con}$ can be obtained by:
\begin{equation}
\label{E16}
\mathcal{L}_{con} = MSE(\mathbf{att}_{m},\psi(\mathbf{att}_{r})) + MSE(\mathbf{att}_{r},\psi(\mathbf{att}_{m})),
\end{equation}
where $\psi(\cdot)$ represents a function that truncates the gradient of the input, and MSE$(\cdot,\cdot)$ denotes the Mean Squared Error loss.

The consistency constraint loss can encourage $\mathbf{att}_{m}$ trained by TSM and $\mathbf{att}_{r}$ trained by VLC to focus on the same action area within the video.
In this way, the localization errors which are caused by the excessive attention of the matching strategy on the most relevant segments can be alleviated.
Besides, the information of the action label text can be transmitted from the video-text language completion model to the WTAL model through the attention mechanism indirectly.

\subsection{Model Training and Inference}
\label{Inference}
\noindent \textbf{Optimizing Process.}
Considering all the aforementioned objectives, our final objective function of the whole framework arrives at: 
\begin{equation}
\label{E17}
\mathcal{L} = \mathcal{L}_{mil} + \alpha \mathcal{L}_{rec} + \beta \mathcal{L}_{c} + \lambda \mathcal{L}_{con},
\end{equation}
where $\alpha, \beta, \lambda$ are the hyper-parameters to balance these four loss terms. 

\noindent \textbf{Model Inference.}
In the test stage, we follow the process of \cite{islam2021hybrid,hong2021cross}.
Firstly, we select those classes with video-level category scores above a threshold for generating proposals.
Then for the selected action classes, we obtain the class-agnostic action proposals by thresholding the attention weights and selecting the continuous components of the remaining segments.
The obtained $i$-th candidate action proposal can be denoted as $\{c_i, s_i, e_i, {conf_i}\}$. 
\begin{table}[!tbp]
 
    \centering
    \setlength{\abovecaptionskip}{-0.1cm}
    \caption{Experimental results of different methods in THUMOS14 dataset.}
    \begin{center}
    \begin{tabular}{p{2.8cm}<{\raggedright}|p{0.3cm}<{\raggedright}p{0.3cm}<{\raggedright}p{0.3cm}<{\raggedright}p{0.3cm}<{\raggedright}p{0.5cm}<{\raggedright}|p{0.7cm}<{\raggedright}}
    \hline\hline
     \multirow{2}{*}{Method} & \multicolumn{5}{c|}{mAP@IoU(\%)} & {Avg} \\ 
    \cline{2-7}
      ~&0.3&0.4&0.5&0.6&0.7&0.3:0.7\\
      \hline\hline
      BasNet(2020)\cite{lee2019background}&44.6&36.0&27.0&18.6&10.4&25.2\\
      RPN(2020)\cite{huang2020relational}&48.2&37.2&27.9&16.7&8.1&27.6\\
      TSCN(2020)\cite{Zhai2020TwoStreamConsensus}&47.8&37.7&28.7&19.4&10.2&28.8\\
      HamNet(2021)\cite{islam2021hybrid}&50.3&41.1&31.0&20.7&11.1&30.8\\
      UGCT(2021)\cite{yang2021uncertainty}&55.5&46.5&35.9&23.8&11.4&34.6\\
      CO2Net(2021)\cite{hong2021cross}&54.5&45.7&38.3&26.4&13.4&35.6\\
      FACNet(2021)\cite{huang2021foreground}&52.6&44.3&33.4&22.5&12.7&33.1\\
      FTCL(2022)\cite{gao2022fine}&55.2&45.2&35.6&23.7&12.2&33.4\\
      ASMLoc(2022)\cite{he2022asm}&\textbf{57.1}&46.8&36.6&25.2&13.1&34.4\\
      DCC(2022)\cite{li2022exploring}&55.9&45.9&35.7&24.3&13.7&35.1\\
      RSKP(2022)\cite{huang2022weakly}&55.8&47.5&38.2&25.4&12.5&35.9\\
      \hline
      Ours&56.2&\textbf{47.8}&\textbf{39.3}&\textbf{27.5}&\textbf{15.2}&\textbf{37.2}\\
      \hline\hline
    \end{tabular}
    \end{center}
    \label{tab1}
\vspace{-0.9cm}
\end{table}
For the confidence score $conf_i$, we follow the AutoLoc \cite{shou2018autoloc} to calculate the outer-inner score of each action proposal through $\bar{\mathbf{S}}$.
Finally, we remove the overlapping
proposals using soft non-maximum suppression.

\section{Experiments}
\subsection{Datasets}
\noindent \textbf{THUMOS14.} THUMOS14 \cite{THUMOS14} dataset contains 200 validation videos and 213 test videos.
There are a totally of 20 categories in the dataset, with an average of 15.5 actions per video. 
Following the same setting as \cite{paul2018w,nguyen2018weakly,narayan20193c,huang2022multi}, we adopt 200 validation videos for training and 213 test videos for testing.

\noindent \textbf{ActivityNet.} ActivityNet \cite{caba2015activitynet} dataset offers a larger benchmark for temporal action localization. There are 10,024 training videos, 4,926 validation videos, and 5,044 testing videos with 200 action categories. Following the experimental setting in \cite{huang2022multi,huang2021foreground,yang2021uncertainty,huang2022weakly}, we adopt all the training videos to train our model and evaluate our proposed method in all the testing videos.

\subsection{Implementation Details}
\noindent \textbf{Evaluation Metrics.} 
We evaluate the proposed method for action localization using mean Average Precision (mAP). The prediction proposal is considered as correct if its action category is predicted correctly and overlaps significantly with the ground truth segment (based on the IoU threshold). We adopt the official evaluation code of ActivityNet to evaluate our method~\cite{caba2015activitynet}.

\begin{table}[!tbp]
    \centering
    \setlength{\abovecaptionskip}{-0.1cm}
    \caption{Experimental results of different methods in ActivityNet1.3 dataset.}
    \begin{center}
    \begin{tabular}{p{2.8cm}<{\raggedright}|p{0.6cm}<{\centering}p{0.6cm}<{\centering}p{0.6cm}<{\centering}|p{1cm}<{\centering}}
    \hline\hline
     \multirow{2}{*}{Method} & \multicolumn{3}{c|}{mAP@IoU(\%)} & {Avg} \\ 
    \cline{2-5}
      ~&0.5&0.75&0.95&0.5:0.95\\
      \hline\hline
      BasNet(2020)\cite{lee2019background}&34.5&22.5&4.9&22.2\\
      TSCN(2020)\cite{Zhai2020TwoStreamConsensus}&25.3&21.4&5.3&21.7\\
      UGCT(2021)\cite{yang2021uncertainty}&39.1&22.4&5.8&23.8\\
      FACNet(2021)\cite{huang2021foreground}&37.6&24.2&6.0&24.0\\
      ACMNet(2021) & 40.1&24.2&6.2&24.6\\
      FTCL(2022)\cite{gao2022fine}&40.0&24.3&\textbf{6.4}&24.8\\
      ASMLoc(2022)\cite{he2022asm}&41.0&24.9&6.2&25.1\\
      DCC(2022)\cite{li2022exploring}&38.8&24.2&5.7&24.3\\
      RSKP(2022)\cite{huang2022weakly}&40.6&24.5&5.9&25.0\\
      \hline
      Ours&\textbf{41.8}&\textbf{26.0}&6.0&\textbf{26.0}\\
      \hline\hline

    \end{tabular}
    \end{center}
    \label{tab2}
 \vspace{-0.9cm}
\end{table}

\noindent \textbf{Feature Extractor.}
Following previous work \cite{paul2018w,narayan20193c,ding2021kfc,2020Adversarial}, the optical flow maps are generated by using the TV-L1 algorithm \cite{inbook}, and we use I3D network \cite{DBLP:conf/cvpr/CarreiraZ17} pre-trained on the Kinetics dataset \cite{article2017} to extract both RGB and optical flow features without fine-turning.

\noindent \textbf{Training Settings.}
we use Adam \cite{kingma2014adam} with a learning rate of 0.0005 and weight decay of 0.001 to optimize our model for about 5,000 iterations on THUMOS14. For ActivityNet1.3, the learning rate is 0.00003 to optimize our model for about 50,000 iterations. For the hyper-parameters in $L_c$, we set $\gamma_1$ as 0.1 and $\gamma_2$ as 0.2. Besides, for the hyper-parameter $\alpha,\beta,\lambda$, we set it as 1.0,1.0,1.5 on THUMOS14 and 1.0,1.0, 0.25 on ActivityNet1.3, respectively.
Our model is implemented by PyTorch 1.8 and trained under Ubuntu 18.04 platform. Hyper-parameter sensitivity analysis can be found in the supplementary materials.

\subsection{Comparison with the State-of-the-Arts
}
We compare the proposed method with state-of-the-art weakly-supervised methods in this section. The results are shown in Table \ref{tab1} and Table \ref{tab2}.
For THUMOS14 datasets, the proposed framework evidently outperforms current state-of-the-art WTAL approaches, especially in high IoU experimental settings. 
On the important criterion: average mAP (0.3:0.7), we surpass the state-of-the-art method \cite{huang2022weakly} by 1.3\%, even surpassing some fully-supervised methods.
For the larger ActivityNet1.3 dataset, our method still obtains 1.0\%mAP improvement over existing the state-of-the-art weakly-supervised methods \cite{huang2022weakly} on average.

\subsection{Ablation Study}
\noindent \textbf{Effectiveness of each component.} 
The proposed framework mainly contains three ingredients: (1) the text-segment mining (TSM) module to replace the existing WTAL model that only uses video information; (2)
Additional video-text language complements (VLC) are used to constrain WTAL models in a self-supervised manner, denoted as $\mathcal{L}_{rec} + \mathcal{L}_{con}$; (3) contrastive loss in the video-text language complements model, denoted as $\mathcal{L}_{c}$.

To verify the effectiveness of each component in the proposed framework, we conduct a comprehensive ablation study to analyze different components in Table \ref{tab3}.
Specifically, we implement four variants of the proposed method as follows: (1) ``Baseline'': Using a convolution layer as a classifier instead of video-text matching in TSM, and only using video information to train the WTAL model;
(2) ``Baseline + $\mathcal{L}_{rec} + \mathcal{L}_{con}$'': Additional video text language complements (VLC) is used to constrain baseline WTAL models in a self-supervised manner;
(3) ``Baseline + $\mathcal{L}_{rec} + \mathcal{L}_{con} +\mathcal{L}_{c}$'': Using contrastive loss in the video-text language complements model; (4) ``TSM+ $\mathcal{L}_{rec} + \mathcal{L}_{con}+\mathcal{L}_{c}$'' : The final framework, replacing the baseline WTAL model with the proposed TSM on the basis of (3);
\vspace{-0.2cm}
\begin{table}[!htbp]
    \centering
    \setlength{\abovecaptionskip}{-0.1cm}
    \caption{Effectiveness of each component on THUMOS14 datasets.}
    \begin{center}
    \begin{tabular}{p{4.2cm}<{\raggedright}|p{0.3cm}<{\raggedright}p{0.3cm}<{\raggedright}p{0.5cm}<{\raggedright}|p{0.7cm}<{\raggedright}}
    \hline\hline
     \multirow{2}{*}{Method} & \multicolumn{3}{c|}{mAP@IoU(\%)} & \multirow{2}{*}{Avg} \\ 
    \cline{2-4}
      ~&0.3&0.5&0.7&~\\
      \hline\hline
      Baseline & 54.5  & 36.5 & 13.0 &34.9\\
      \hline
      Baseline + $\mathcal{L}_{rec} + \mathcal{L}_{con}$ & 55.0  & 37.8  & 14.0 &35.9 \\
      Baseline + $\mathcal{L}_{rec} + \mathcal{L}_{con}+ L_{c}$ & 55.7 & 38.3 & 13.8 & 36.3 \\
      TSM + $\mathcal{L}_{rec} + \mathcal{L}_{con} + L_{c}$ & 56.2  & 39.3  &15.2 &37.2 \\
       \hline\hline
    \end{tabular}
    \end{center}
    \label{tab3}
\vspace{-0.6cm}
\end{table}

By comparing the performance of methods ``TSM + $\mathcal{L}_{rec} + \mathcal{L}_{con} + \mathcal{L}_{c}$'' and ``Baseline + $\mathcal{L}_{rec} + \mathcal{L}_{con} + \mathcal{L}_{c}$'', we can conclude that the text-segment mining is better than generally WTAL model using only convolution classifier without action label text information, which brings about 0.9\% performance improvement on THUMOS14 dataset.
When we ablate the measurement of learning loss $\mathcal{L}_c$ and the additional video-text language completion model $\mathcal{L}_{rec} + \mathcal{L}_{con}$ step-by-step, the performance under all the experiment settings could be gradually decreased.
To be specific, by comparing the methods ``Baseline + $\mathcal{L}_{rec} + \mathcal{L}_{con} $'' with ``Baseline'', we can conclude that the proposed video-text language model can constrain the WTAL model by self-supervision and indirectly transfer text information to it, which brings about 1.0\% mAP performance improvement on THUMOS14 dataset.
Besides, comparing the methods ``Baseline + $\mathcal{L}_{rec} + \mathcal{L}_{con}+\mathcal{L}_{c}$'' with ``Baseline + $\mathcal{L}_{rec} + \mathcal{L}_{con}$'', we can also verify the effectiveness of the contrastive loss in the VLC. 

Futhermore, we evaluated the frame-level classification results on THUMOS14.
Compared with the baseline, after using the TSM model, the false positive rate (FPR) dropped from 26.0\% to 23.8\%, and after using the VLC model, the false negative rate
(FNR) decreased from 28.0\% to 26.9\%. This shows that TSM can effectively alleviate the problem that the background segment is misclassified as a groundtruth action, thus effectively alleviating the over-complete
problem while VLC can effectively alleviate the problem that the groundtruth action segment is misclassified as background, thus
effectively alleviating the incomplete problem. 

\noindent \textbf{Comparisons with different prompts in text-segment mining model.}
We compare the effects of handcraft prompts ``a video of [CLS]'' and the learnable prompt on the text-segment mining models in Table \ref{tab4}.
Compared with the handcraft prompt in the text-segment mining model, the learnable prompt achieves better performance.
It is because, by making it learnable, textual contexts can achieve better transferability in downstream video-text matching tasks by directly optimizing the contexts using back-propagation. 
%
\vspace{-0.2cm}
\begin{table}[!htbp]
    \vspace{-0.1cm}
    \centering
    \setlength{\abovecaptionskip}{-0.1cm}
    \caption{Comparisons with different prompts in classification model on THUMOS14 dataset.}
    \begin{center}
    \begin{tabular}{p{2.8cm}<{\raggedright}|p{0.35cm}<{\raggedright}p{0.35cm}<{\raggedright}p{0.5cm}<{\raggedright}|p{0.4cm}<{\raggedright}}
    \hline\hline
     \multirow{2}{*}{Method} & \multicolumn{3}{c|}{mAP@IoU(\%)} & \multirow{2}{*}{Avg} \\ 
    \cline{2-4}
      ~&0.3&0.5&0.7&~\\
      \hline\hline
      handcraft prompt&55.1&38.1&14.1&36.1\\
      learnable prompt&\textbf{56.2}&\textbf{39.3}&\textbf{15.2}&\textbf{37.2}\\
      \hline\hline
    \end{tabular}
    \end{center}
    \label{tab4}
\vspace{-0.7cm}
\end{table}

\noindent \textbf{Comparisons with different types of consistency constraint loss.}
We also evaluate the effect of different types of consistency constraints. 
Specifically, we implement five variants of the constraints on the VLC and TSM  model in different ways: (1) ``w/o $\mathcal{L}_{con}$'': The VLC model is not used, and only TSM is used as the baseline;
(2) ``Share'': $\mathcal{L}_{con}$ is not used, but the VLC and TSM share the parameters of the attention module; (3) ``KL'': Using Kullback Leibler divergence~\cite{2013Kullback} as loss function $\mathcal{L}_{con}$; (4)``MAE'': Using Mean Absolute Error as loss $\mathcal{L}_{con}$; (5)``MSE'': Using Mean Square Error as loss $\mathcal{L}_{con}$.

The result in Table \ref{tab42} shows that using an additional video-text completion model to constrain the WTAL model can effectively improve localization performance, and using MSE as the consistency constraint loss is more suitable. 
\vspace{-0.1cm}
\begin{table}[!htbp]
    \vspace{-0.1cm}
    \centering
    \setlength{\abovecaptionskip}{-0.1cm}
    \caption{Comparisons with types of consistency constraint loss on THUMOS14 dataset.}
    \begin{center}
    \begin{tabular}{p{1.8cm}<{\raggedright}|p{0.8cm}<{\raggedright}p{0.8cm}<{\raggedright}p{0.8cm}<{\raggedright}p{0.8cm}<{\raggedright}p{0.8cm}<{\raggedright}}
    \hline\hline
     {Method} & w/o & Share & KL & MAE & MSE \\ 
    \hline
     Avg mAP&35.4&35.8&36.9&36.4&\textbf{37.2}\\
      \hline\hline
      
    \end{tabular}
    \end{center}
    \label{tab42}
\vspace{-0.7cm}
\end{table}

\noindent \textbf{Comparisons with different types of language reconstructor in the video-text language completion model.} 

We compare the performance impact of using different prompt templates to generate action descriptions in the completion model.

To verify the effectiveness of additional video-text language completion model, we compare the effects of different types of language reconstructor on the localization result. 
Specifically, we compared three different reconstructors, Transformer, GRU \cite{cho2014learning} and LSTM \cite{hochreiter1997long} in Table \ref{tab43}. In addition, ``w/o'' represents only the TSM model used.
As shown in Table \ref{tab43}, we can conclude that no matter which language reconstructor is used, the video-text language completion model could improve the performance of the proposed framework, by imposing self-supervised constraints on TSM.
Besides, we can conclude that the Transformer structure is more suitable to be used as a language reconstructor in our framework.
\vspace{-0.1cm}
\begin{table}[!htbp]
 \vspace{-0.1cm}
    \centering
    \setlength{\abovecaptionskip}{-0.1cm}
    \caption{Comparisons with different types of language reconstructors in the video-text language completion model on THUMOS14 dataset.}
    \begin{center}
 \begin{tabular}{p{1.8cm}<{\raggedright}|p{0.4cm}<{\centering}p{0.5cm}<{\centering}p{0.6cm}<{\centering}p{2cm}<{\centering}}
    \hline\hline
     {Method} & w/o & GRU & LSTM & Transformer \\ 
    \hline
     Avg mAP&35.4&36.7&36.1&\textbf{37.2}\\
      \hline\hline
    \end{tabular}
    \end{center}
    \label{tab43}
\vspace{-0.7cm}
\end{table}

\noindent \textbf{Comparison of action descriptions generated by different prompt templates in language completion model.}
We compared the performance influence of action descriptions generated by different prompt templates in the language completion model in Table \ref{tab44}. The results of all types of prompt templates can outperform existing state-of-the-art results, as shown in Table \ref{tab1}. These results indicate that it is necessary to use video-text language completion model to constrain WTAL model.

\vspace{-0.2cm}
\begin{table}[!htbp]
  \vspace{-0.1cm}
    \centering
    \setlength{\abovecaptionskip}{-0.1cm}
    \caption{Comparisons with different prompts in completion model on THUMOS14 dataset.}
    \vspace{-0.1cm}
    \begin{center}
    \begin{tabular}{p{3.5cm}<{\raggedright}|p{0.35cm}<{\raggedright}p{0.35cm}<{\raggedright}p{0.5cm}<{\raggedright}|p{0.4cm}<{\raggedright}}
    \hline\hline
     \multirow{2}{*}{Prompt} & \multicolumn{3}{c|}{mAP@IoU(\%)} & \multirow{2}{*}{Avg} \\ 
    \cline{2-4}
      ~&0.3&0.5&0.7&~\\
      \hline\hline
      a [CLS] &55.9&38.7&14.6&36.8\\
      a video of action [CLS] &55.1&38.5&15.0&36.6\\
      a video of the [CLS]  &\textbf{56.2}&\textbf{39.3}&\textbf{15.2}&\textbf{37.2}\\
      \hline\hline
    \end{tabular}
    \end{center}
    \label{tab44}
\vspace{-0.7cm}
\end{table}

\noindent \textbf{Integrating our framework to existing methods.}
The proposed method can be easily extended to existing WTAL models and improve their performance.
To verify the scalability of the proposed framework, we design three sets of experiments to extend the proposed framework to existing methods: (1) ``+TSM'': Using the proposed TSM to replace convolution classifier of existing WTAL model; (2) ``+VLC'': Additional VLC model are used to constrain WTAL models in a self-supervised manner; (3) ``+TSM+VLC'': extended all components of our framework to existing WTAL model.
As shown in Table \ref{tab6}, we can clearly conclude that both of the proposed TSM and VLC can greatly improve the performance of two existing methods, verifying the effectiveness of leveraging action label text information to expand WTAL model.
\vspace{-0.2cm}
\begin{table}[!htbp]
 
    \centering
    \setlength{\abovecaptionskip}{-0.1cm}
    \caption{Integrating our framework to existing methods on THUMOS14 dataset.}
    \vspace{-0.1cm}
    \begin{center}
    \begin{tabular}{p{4cm}<{\raggedright}|p{0.4cm}<{\raggedright}p{0.4cm}<{\raggedright}p{0.5cm}<{\raggedright}|p{0.6cm}<{\raggedright}}
    \hline\hline
     \multirow{2}{*}{Method} & \multicolumn{3}{c|}{mAP@IoU(\%)} & \multirow{2}{*}{Avg} \\ 
    \cline{2-4}
      ~&0.3&0.5&0.7&~\\
      \hline\hline
      BaSNet \cite{lee2019background}&44.6&27.0&10.4&27.3\\
      BaSNet + TSM &48.2&31.7&9.7&29.5\\
      BaSNet + VLC &48.6 & 32.0 & 10.3&30.2\\
      BaSNet + TSM + VLC & \textbf{49.0} & \textbf{32.5} & \textbf{10.7} &\textbf{30.6} \\
      \hline
      HAMNet \cite{islam2021hybrid}& 50.3 & 31.0&11.1 &30.8\\
      HAMNet + TSM &51.8&34.7&11.8&32.7\\
      HAMNet + VLC &51.5 & 36.0 & 12.8&33.6\\
      HAMNet + TSM + VLC & \textbf{52.3} & \textbf{37.4} & \textbf{13.4} & \textbf{34.5}\\

      \hline\hline
    \end{tabular}
    \end{center}
    \label{tab6}
\vspace{-0.8cm}
\end{table}

\subsection{Qualitative Analysis}

\begin{figure}[!htbp]
\begin{center}
    \includegraphics[width=1.0\linewidth]{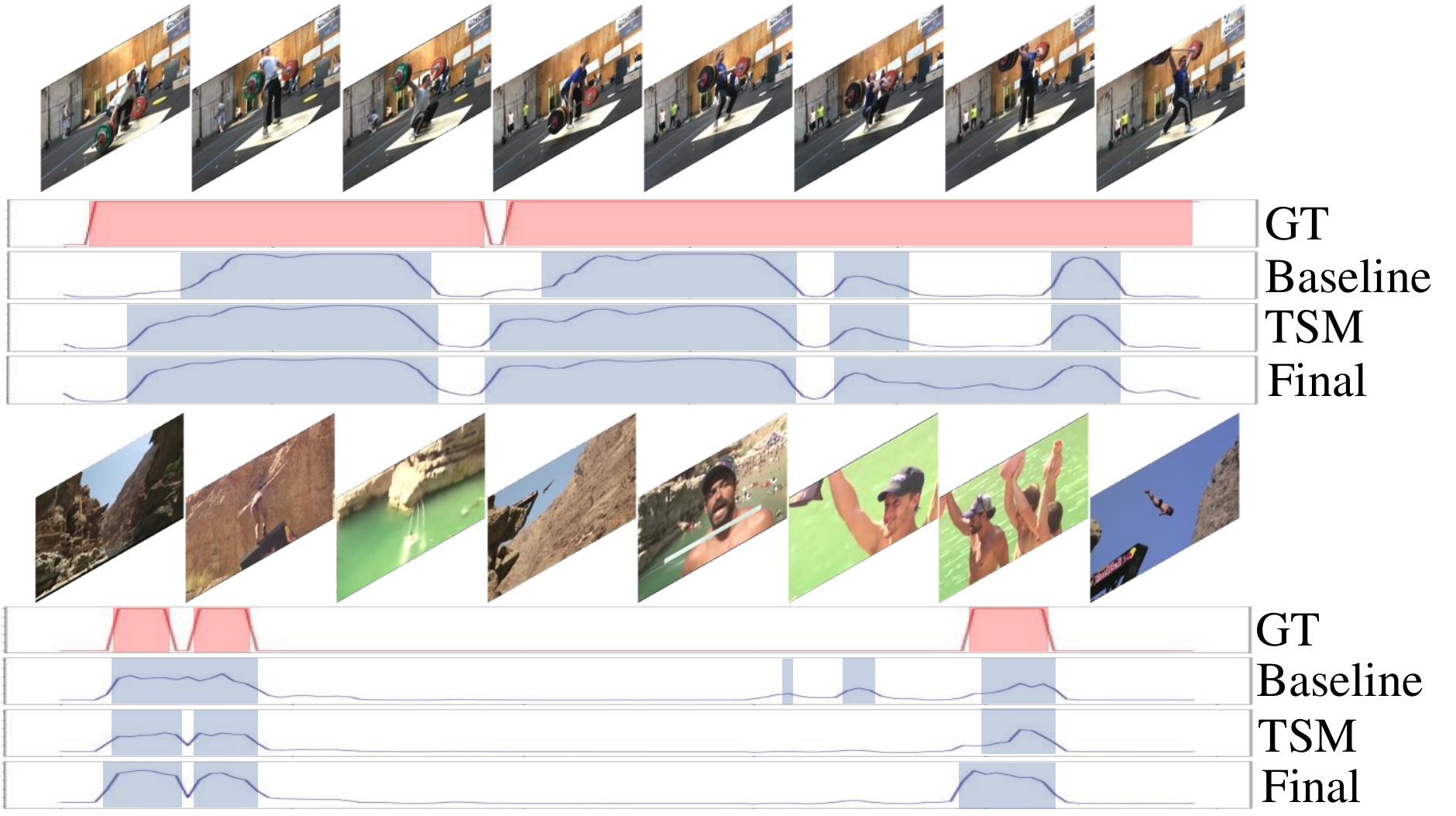}
\end{center}
 \vspace{-0.2cm}
   \caption{{Two prediction examples on THUMOS14 dataset. }}
\label{fig:Qua}

\end{figure}

We visualize some examples of the detected action instances in Figure \ref{fig:Qua}. For each example, the top line represents the segment of the video, the following four lines in order are the ground truth of the action in the video, the localization results generated by the baseline model, the localization results generated by the text-segment mining, and the localization results generated by our final framework. As can be seen from this figure, introducing the text information in the category annotation into the WTAL model in both direct and indirect ways, helps to generate more accurate 
localization results and suppress the response of background fragments to a certain extent.
\label{subsubsec: Qualitative}
\vspace{-0.7cm}
\section{Conclusion}
We introduce a new framework to leverage the text information to boost WTAL from two aspects, \ie text-segment mining, and video-text language completion. With the help of text information, the proposed method can focus on the action-category-related areas in the video and improve the performance of WTAL tasks. Extensive experiments demonstrate that the proposed method achieves state-of-the-art performances on two popular datasets, and both of the proposed objectives can be directly extended to the existing WTAL methods to improve their performances.

\noindent \textbf{Limitation.}
One major limitation in this work is that we must train the text-segment mining and video- text language completion models at the same time, resulting in the model size being twice as the original size. In the future, We will explore more efficient manners to make full use of text information in tags to boost WTAL.

\noindent
\textbf{Acknowledgments:} 
This work was supported in part by the National Key Research and Development Program of China under Grant 2018AAA0103202; in part by the National Natural Science Foundation of China under Grants 62036007, U22A2096 and 62176198; in part by the Technology Innovation Leading Program of Shaanxi under Grant 2022QFY01-15; in part by Open Research Projects of Zhejiang Lab under Grant 2021KG0AB01.

{\small
\bibliographystyle{ieee_fullname}
\bibliography{egbib}

\begin{thebibliography}{10}\itemsep=-1pt

\bibitem{caba2015activitynet}
Fabian Caba~Heilbron, Victor Escorcia, Bernard Ghanem, and Juan Carlos~Niebles.
\newblock Activitynet: A large-scale video benchmark for human activity
  understanding.
\newblock In {\em Proceedings of the ieee conference on computer vision and
  pattern recognition}, pages 961--970, 2015.

\bibitem{DBLP:conf/cvpr/CarreiraZ17}
Jo{\~{a}}o Carreira and Andrew Zisserman.
\newblock Quo vadis, action recognition? {A} new model and the kinetics
  dataset.
\newblock In {\em 2017 {IEEE} Conference on Computer Vision and Pattern
  Recognition, {CVPR} 2017, Honolulu, HI, USA, July 21-26, 2017}, pages
  4724--4733. {IEEE} Computer Society, 2017.

\bibitem{cho2014learning}
Kyunghyun Cho, Bart Van~Merri{\"e}nboer, Caglar Gulcehre, Dzmitry Bahdanau,
  Fethi Bougares, Holger Schwenk, and Yoshua Bengio.
\newblock Learning phrase representations using rnn encoder-decoder for
  statistical machine translation.
\newblock {\em arXiv preprint arXiv:1406.1078}, 2014.

\bibitem{ding2021kfc}
Xinpeng Ding, Nannan Wang, Xinbo Gao, Jie Li, Xiaoyu Wang, and Tongliang Liu.
\newblock Kfc: An efficient framework for semi-supervised temporal action
  localization.
\newblock {\em IEEE Transactions on Image Processing}, 30:6869--6878, 2021.

\bibitem{ding2021crnet}
Xinpeng Ding, Nannan Wang, Jie Li, and Xinbo Gao.
\newblock Crnet: Centroid radiation network for temporal action localization.
\newblock In {\em Chinese Conference on Pattern Recognition and Computer Vision
  (PRCV)}, pages 29--41. Springer, 2021.

\bibitem{Ding_2021_ICCV}
Xinpeng Ding, Nannan Wang, Shiwei Zhang, De Cheng, Xiaomeng Li, Ziyuan Huang,
  Mingqian Tang, and Xinbo Gao.
\newblock Support-set based cross-supervision for video grounding.
\newblock In {\em Proceedings of the IEEE/CVF International Conference on
  Computer Vision (ICCV)}, pages 11573--11582, October 2021.

\bibitem{gao2022fine}
Junyu Gao, Mengyuan Chen, and Changsheng Xu.
\newblock Fine-grained temporal contrastive learning for weakly-supervised
  temporal action localization.
\newblock {\em arXiv preprint arXiv:2203.16800}, 2022.

\bibitem{gongself}
Guoqiang Gong, Liangfeng Zheng, Wenhao Jiang, and Yadong Mu.
\newblock Self-supervised video action localization with adversarial temporal
  transforms.

\bibitem{he2022asm}
Bo He, Xitong Yang, Le Kang, Zhiyu Cheng, Xin Zhou, and Abhinav Shrivastava.
\newblock Asm-loc: Action-aware segment modeling for weakly-supervised temporal
  action localization.
\newblock {\em arXiv preprint arXiv:2203.15187}, 2022.

\bibitem{hochreiter1997long}
Sepp Hochreiter and J{\"u}rgen Schmidhuber.
\newblock Long short-term memory.
\newblock {\em Neural computation}, 9(8):1735--1780, 1997.

\bibitem{hong2021cross}
Fa-Ting Hong, Jia-Chang Feng, Dan Xu, Ying Shan, and Wei-Shi Zheng.
\newblock Cross-modal consensus network for weakly supervised temporal action
  localization.
\newblock In {\em Proceedings of the 29th ACM International Conference on
  Multimedia}, pages 1591--1599, 2021.

\bibitem{huang2020relational}
Linjiang Huang, Yan Huang, Wanli Ouyang, Liang Wang, et~al.
\newblock Relational prototypical network for weakly supervised temporal action
  localization.
\newblock 2020.

\bibitem{huang2021foreground}
Linjiang Huang, Liang Wang, and Hongsheng Li.
\newblock Foreground-action consistency network for weakly supervised temporal
  action localization.
\newblock In {\em Proceedings of the IEEE/CVF International Conference on
  Computer Vision}, pages 8002--8011, 2021.

\bibitem{huang2022multi}
Linjiang Huang, Liang Wang, and Hongsheng Li.
\newblock Multi-modality self-distillation for weakly supervised temporal
  action localization.
\newblock {\em IEEE Transactions on Image Processing}, 2022.

\bibitem{huang2022weakly}
Linjiang Huang, Liang Wang, and Hongsheng Li.
\newblock Weakly supervised temporal action localization via representative
  snippet knowledge propagation.
\newblock {\em arXiv preprint arXiv:2203.02925}, 2022.

\bibitem{islam2021hybrid}
Ashraful Islam, Chengjiang Long, and Richard~J. Radke.
\newblock A hybrid attention mechanism for weakly-supervised temporal action
  localization, 2021.

\bibitem{THUMOS14}
Y.-G. Jiang, J. Liu, A. Roshan~Zamir, G. Toderici, I. Laptev, M. Shah, and R.
  Sukthankar.
\newblock {THUMOS} challenge: Action recognition with a large number of
  classes.
\newblock \url{http://crcv.ucf.edu/THUMOS14/}, 2014.

\bibitem{ju2021prompting}
Chen Ju, Tengda Han, Kunhao Zheng, Ya Zhang, and Weidi Xie.
\newblock Prompting visual-language models for efficient video understanding.
\newblock {\em arXiv preprint arXiv:2112.04478}, 2021.

\bibitem{article2017}
Will Kay, J. Carreira, Karen Simonyan, Brian Zhang, Chloe Hillier, Sudheendra
  Vijayanarasimhan, Fabio Viola, Tim Green, Trevor Back, Paul Natsev, Mustafa
  Suleyman, and Andrew Zisserman.
\newblock The kinetics human action video dataset.
\newblock 05 2017.

\bibitem{kingma2014adam}
Diederik~P Kingma and Jimmy Ba.
\newblock Adam: A method for stochastic optimization.
\newblock {\em arXiv preprint arXiv:1412.6980}, 2014.

\bibitem{lee2021cross}
Jun-Tae Lee, Mihir Jain, Hyoungwoo Park, and Sungrack Yun.
\newblock Cross-attentional audio-visual fusion for weakly-supervised action
  localization.
\newblock In {\em International Conference on Learning Representations}, 2021.

\bibitem{lee2019background}
Pilhyeon Lee, Youngjung Uh, and Hyeran Byun.
\newblock Background suppression network for weakly-supervised temporal action
  localization.
\newblock {\em arXiv preprint arXiv:1911.09963}, 2019.

\bibitem{lee2021weakly}
Pilhyeon Lee, Jinglu Wang, Yan Lu, and Hyeran Byun.
\newblock Weakly-supervised temporal action localization by uncertainty
  modeling.
\newblock In {\em Proceedings of the AAAI Conference on Artificial
  Intelligence}, volume~35, pages 1854--1862, 2021.

\bibitem{lei2019tvqa+}
Jie Lei, Licheng Yu, Tamara~L Berg, and Mohit Bansal.
\newblock Tvqa+: Spatio-temporal grounding for video question answering.
\newblock {\em arXiv preprint arXiv:1904.11574}, 2019.

\bibitem{li2022exploring}
Jingjing Li, Tianyu Yang, Wei Ji, Jue Wang, and Li Cheng.
\newblock Exploring denoised cross-video contrast for weakly-supervised
  temporal action localization.
\newblock In {\em Proceedings of the IEEE/CVF Conference on Computer Vision and
  Pattern Recognition}, pages 19914--19924, 2022.

\bibitem{Lin_2019_ICCV}
Tianwei Lin, Xiao Liu, Xin Li, Errui Ding, and Shilei Wen.
\newblock Bmn: Boundary-matching network for temporal action proposal
  generation.
\newblock In {\em Proceedings of the IEEE/CVF International Conference on
  Computer Vision (ICCV)}, October 2019.

\bibitem{lin2020weakly}
Zhijie Lin, Zhou Zhao, Zhu Zhang, Qi Wang, and Huasheng Liu.
\newblock Weakly-supervised video moment retrieval via semantic completion
  network.
\newblock In {\em Proceedings of the AAAI Conference on Artificial
  Intelligence}, volume~34, pages 11539--11546, 2020.

\bibitem{2020WeaklyEM}
Zhekun Luo, Devin Guillory, Baifeng Shi, Wei Ke, and Huijuan Xu.
\newblock Weakly-supervised action localization with expectation-maximization
  multi-instance learning.
\newblock 2020.

\bibitem{min2020adversarial}
Kyle Min and Jason~J. Corso.
\newblock Adversarial background-aware loss for weakly-supervised temporal
  activity localization, 2020.

\bibitem{2020Adversarial}
Kyle Min and Jason~J. Corso.
\newblock Adversarial background-aware loss for weakly-supervised temporal
  activity localization.
\newblock 2020.

\bibitem{mun2020local}
Jonghwan Mun, Minsu Cho, and Bohyung Han.
\newblock Local-global video-text interactions for temporal grounding.
\newblock In {\em Proceedings of the IEEE/CVF Conference on Computer Vision and
  Pattern Recognition}, pages 10810--10819, 2020.

\bibitem{narayan2021d2}
Sanath Narayan, Hisham Cholakkal, Munawar Hayat, Fahad~Shahbaz Khan, Ming-Hsuan
  Yang, and Ling Shao.
\newblock D2-net: Weakly-supervised action localization via discriminative
  embeddings and denoised activations.
\newblock In {\em Proceedings of the IEEE/CVF International Conference on
  Computer Vision}, pages 13608--13617, 2021.

\bibitem{narayan20193c}
Sanath Narayan, Hisham Cholakkal, Fahad~Shahbaz Khan, and Ling Shao.
\newblock 3c-net: Category count and center loss for weakly-supervised action
  localization.
\newblock In {\em Proceedings of the IEEE International Conference on Computer
  Vision}, pages 8679--8687, 2019.

\bibitem{nguyen2018weakly}
Phuc Nguyen, Ting Liu, Gautam Prasad, and Bohyung Han.
\newblock Weakly supervised action localization by sparse temporal pooling
  network.
\newblock In {\em Proceedings of the IEEE Conference on Computer Vision and
  Pattern Recognition}, pages 6752--6761, 2018.

\bibitem{paul2018w}
Sujoy Paul, Sourya Roy, and Amit~K Roy-Chowdhury.
\newblock W-talc: Weakly-supervised temporal activity localization and
  classification.
\newblock In {\em Proceedings of the European Conference on Computer Vision
  (ECCV)}, pages 563--579, 2018.

\bibitem{pennington2014glove}
Jeffrey Pennington, Richard Socher, and Christopher~D Manning.
\newblock Glove: Global vectors for word representation.
\newblock In {\em Proceedings of the 2014 conference on empirical methods in
  natural language processing (EMNLP)}, pages 1532--1543, 2014.

\bibitem{2013Kullback}
Daniel Polani.
\newblock {\em Kullback-Leibler Divergence}.
\newblock Encyclopedia of Systems Biology, 2013.

\bibitem{radford2021learning}
Alec Radford, Jong~Wook Kim, Chris Hallacy, Aditya Ramesh, Gabriel Goh,
  Sandhini Agarwal, Girish Sastry, Amanda Askell, Pamela Mishkin, Jack Clark,
  et~al.
\newblock Learning transferable visual models from natural language
  supervision.
\newblock In {\em International Conference on Machine Learning}, pages
  8748--8763. PMLR, 2021.

\bibitem{shou2018autoloc}
Zheng Shou, Hang Gao, Lei Zhang, Kazuyuki Miyazawa, and Shih-Fu Chang.
\newblock Autoloc: Weakly-supervised temporal action localization in untrimmed
  videos.
\newblock In {\em Proceedings of the European Conference on Computer Vision
  (ECCV)}, pages 154--171, 2018.

\bibitem{singh2017hide}
Krishna~Kumar Singh and Yong~Jae Lee.
\newblock Hide-and-seek: Forcing a network to be meticulous for
  weakly-supervised object and action localization.
\newblock In {\em 2017 IEEE international conference on computer vision
  (ICCV)}, pages 3544--3553. IEEE, 2017.

\bibitem{su2021improving}
Rui Su, Dong Xu, Luping Zhou, and Wanli Ouyang.
\newblock Improving weakly supervised temporal action localization by
  exploiting multi-resolution information in temporal domain.
\newblock {\em IEEE Transactions on Image Processing}, 30:6659--6672, 2021.

\bibitem{tan2021relaxed}
Jing Tan, Jiaqi Tang, Limin Wang, and Gangshan Wu.
\newblock Relaxed transformer decoders for direct action proposal generation.
\newblock {\em arXiv preprint arXiv:2102.01894}, 2021.

\bibitem{turin1960introduction}
George Turin.
\newblock An introduction to matched filters.
\newblock {\em IRE transactions on Information theory}, 6(3):311--329, 1960.

\bibitem{wang2018reconstruction}
Bairui Wang, Lin Ma, Wei Zhang, and Wei Liu.
\newblock Reconstruction network for video captioning.
\newblock In {\em Proceedings of the IEEE conference on computer vision and
  pattern recognition}, pages 7622--7631, 2018.

\bibitem{wang2017untrimmednets}
Limin Wang, Yuanjun Xiong, Dahua Lin, and Luc Van~Gool.
\newblock Untrimmednets for weakly supervised action recognition and detection.
\newblock In {\em Proceedings of the IEEE conference on Computer Vision and
  Pattern Recognition}, pages 4325--4334, 2017.

\bibitem{inbook}
Andreas Wedel, Thomas Pock, Christopher Zach, Horst Bischof, and Daniel
  Cremers.
\newblock {\em An Improved Algorithm for TV-L1 Optical Flow}, pages 23--45.
\newblock 07 2009.

\bibitem{yang2021uncertainty}
Wenfei Yang, Tianzhu Zhang, Xiaoyuan Yu, Tian Qi, Yongdong Zhang, and Feng Wu.
\newblock Uncertainty guided collaborative training for weakly supervised
  temporal action detection.
\newblock In {\em Proceedings of the IEEE/CVF Conference on Computer Vision and
  Pattern Recognition}, pages 53--63, 2021.

\bibitem{zeng2019breaking}
Runhao Zeng, Chuang Gan, Peihao Chen, Wenbing Huang, Qingyao Wu, and Mingkui
  Tan.
\newblock Breaking winner-takes-all: Iterative-winners-out networks for weakly
  supervised temporal action localization.
\newblock {\em IEEE Transactions on Image Processing}, 28(12):5797--5808, 2019.

\bibitem{Zhai2020TwoStreamConsensus}
Yuanhao Zhai, Le Wang, Wei Tang, Qilin Zhang, Junsong Yuan, and Gang Hua.
\newblock Two-stream consensus networks for weakly-supervised temporal action
  localization.
\newblock In {\em 16th European Conference on Computer Vision (ECCV)}, August
  2020.

\bibitem{Zhang_2021_CVPR}
Chuhan Zhang, Ankush Gupta, and Andrew Zisserman.
\newblock Temporal query networks for fine-grained video understanding.
\newblock In {\em Proceedings of the IEEE/CVF Conference on Computer Vision and
  Pattern Recognition (CVPR)}, pages 4486--4496, June 2021.

\bibitem{zhang2022actionformer}
Chenlin Zhang, Jianxin Wu, and Yin Li.
\newblock Actionformer: Localizing moments of actions with transformers.
\newblock {\em arXiv preprint arXiv:2202.07925}, 2022.

\bibitem{zhang2020learning}
Songyang Zhang, Houwen Peng, Jianlong Fu, and Jiebo Luo.
\newblock Learning 2d temporal adjacent networks for moment localization with
  natural language.
\newblock In {\em Proceedings of the AAAI Conference on Artificial
  Intelligence}, volume~34, pages 12870--12877, 2020.

\bibitem{zheng2022weakly}
Minghang Zheng, Yanjie Huang, Qingchao Chen, and Yang Liu.
\newblock Weakly supervised video moment localization with contrastive negative
  sample mining.
\newblock In {\em Proceedings of the AAAI Conference on Artificial
  Intelligence}, volume~1, page~3, 2022.

\bibitem{zhong2018step}
Jia-Xing Zhong, Nannan Li, Weijie Kong, Tao Zhang, Thomas~H Li, and Ge Li.
\newblock Step-by-step erasion, one-by-one collection: a weakly supervised
  temporal action detector.
\newblock In {\em Proceedings of the 26th ACM international conference on
  Multimedia}, pages 35--44, 2018.

\end{thebibliography}
}

\end{document}